\documentclass{article}

\usepackage{microtype}
\usepackage{graphicx}
\usepackage{subcaption}
\usepackage{booktabs} 

\usepackage{hyperref}

\usepackage[accepted]{icml2026}

\usepackage{amsmath}
\usepackage{amssymb}
\usepackage{mathtools}
\usepackage{amsthm}

\usepackage[capitalize,noabbrev]{cleveref}

\usepackage{pifont}
\newcommand{\cmark}{\ding{51}}
\newcommand{\xmark}{\ding{55}}

\usepackage{multirow}
\usepackage{svg}

\theoremstyle{plain}

\theoremstyle{definition}

\theoremstyle{remark}

\usepackage[textsize=tiny]{todonotes}

\icmltitlerunning{MacArena: Benchmarking Computer Use Agents on an Online macOS Environment}

\begin{document}

\twocolumn[
  \icmltitle{MacArena: Benchmarking Computer Use Agents \\ on an Online macOS Environment}




  \begin{icmlauthorlist}
    \icmlauthor{Victor Muryn}{macpaw}
    \icmlauthor{Maksym Shamrai}{macpaw}
    \icmlauthor{Sofiia Mazepa}{macpaw}
    \icmlauthor{Yehor Khodysko}{macpaw}
  \end{icmlauthorlist}

  \icmlaffiliation{macpaw}{MacPaw Research, Kyiv, Ukraine}

  \icmlcorrespondingauthor{Victor Muryn}{victormuryn@macpaw.com}
  \icmlcorrespondingauthor{Maksym Shamrai}{mshamrai@macpaw.com}

  \icmlkeywords{benchmark, computer-use agents, macOS, verifiable reward, ICML}

  \vskip 0.3in
]



\printAffiliationsAndNotice{}  

\begin{abstract}
Computer-use agents (CUAs) operate graphical user interfaces (GUIs) through vision and control primitives, and their capabilities have advanced rapidly, driven in part by standardized online evaluation benchmarks such as OSWorld, which serve both as evaluation tools and as training environments for reinforcement learning. However, macOS remains underserved in this landscape: the only existing benchmark, macOSWorld, covers a narrow slice of first-party applications with simpler tasks, and runs on x86 virtual machines incompatible with Apple Silicon. We introduce MacArena, a benchmark of 421 manually verified tasks spanning 50 applications that combines a curated port of OSWorld tasks, content sourced from macOSWorld, and 49 new macOS-native tasks, all running on Apple's native Virtualization framework on Apple Silicon. We argue that macOS presents distinct GUI challenges beyond what Linux-based benchmarks capture, and our evaluation supports this claim: strong model performance on existing benchmarks can reflect familiarity with task distributions rather than genuine cross-platform GUI competence. Notably, model rankings invert between ported and macOS-native tasks, with a leading model trailing by over 26\% on the MacArena subset, suggesting that macOS poses a genuinely harder environment for current GUI agents\footnote{Code available online \url{https://github.com/MacPaw/MacArena}}.
\end{abstract}

\section{Introduction}

\begin{figure*}[htbp]
    \centering
    \includegraphics[width=\textwidth]{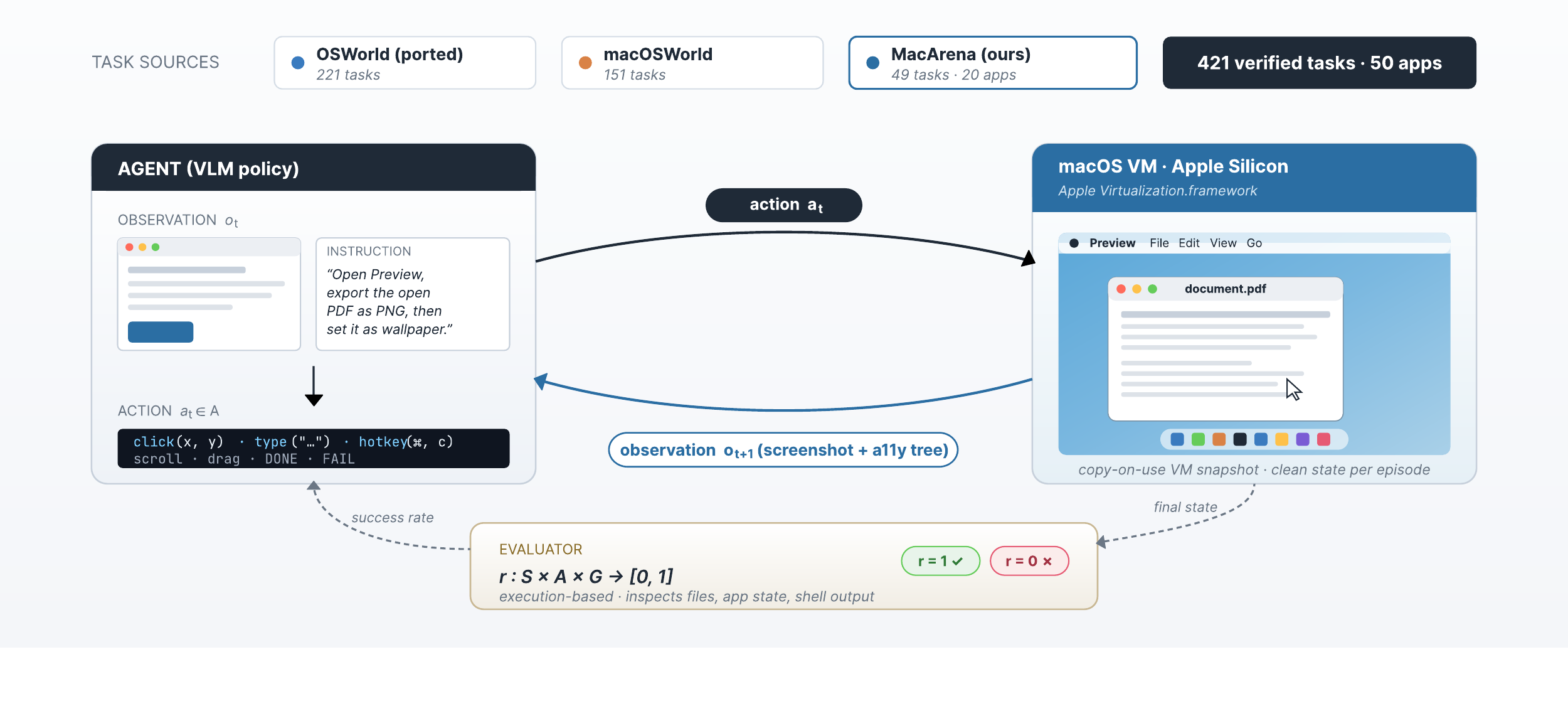}
    \caption{Overview of MacArena. Tasks are drawn from three sources: OSWorld (ported to macOS), macOSWorld, and 49 newly collected macOS-native tasks, totaling 421 human-verified tasks across 50 applications. At each timestep, the agent receives a screenshot and an accessibility tree as observations and produces an action executed within an Apple Silicon VM running via Apple's Virtualization framework. An execution-based evaluator inspects the final environment state to assign a score $r \in [0, 1]$}
    \label{fig:hero}
\end{figure*}

Computer-use agents (CUAs) are systems capable of interacting with graphical user interfaces (GUIs) through direct manipulation — clicking, dragging, and typing on visible on-screen elements~\cite{Sager_2026, nguyen2025guiagentssurvey}. CUAs operate directly on the pixel-level representations that human users see, enabling them to navigate applications, complete multi-step tasks, and respond to dynamic interface states~\cite{zheng2023seeact}. Their ability to generalize across diverse software environments without requiring programmatic access to underlying systems positions them as a promising direction toward general-purpose digital assistants capable of executing real-world computer tasks on behalf of users.

Benchmarking CUA capabilities has been an active research area, with several interactive evaluation environments proposed in recent years. OSWorld~\cite{OSWorld} established the leading cross-platform benchmark, spanning Linux and Windows, using real applications and executable tasks. Yet, macOS remains underserved as an evaluation target. The only existing macOS benchmark, macOSWorld~\cite{macosworld}, covers a narrow slice of the platform's task space: UI navigation sequences are simpler, and task specifications are less ambiguous than those found in cross-platform benchmarks. Coverage is also limited almost exclusively to built-in applications, leaving a wide range of commonly used third-party software unevaluated — a significant gap given how central such software is to real-world macOS usage. Furthermore, macOSWorld relies on x86-based virtual machines, making it incompatible with the entire modern Apple Silicon lineup and unable to reflect the performance characteristics of current hardware. This raises a broader question: whether macOS presents distinct challenges for GUI agents that go beyond what existing Linux-based benchmarks capture.

To address these gaps, we introduce \textbf{MacArena}, a benchmark for evaluating computer-use agents on macOS (Figure~\ref{fig:hero}). MacArena is built from three sources: a curated port of tasks from OSWorld to macOS, a set of tasks sourced from macOSWorld, and 49 novel macOS-specific tasks that increase task complexity and broaden coverage to non-standard applications.

Our main contributions are:
\begin{itemize}
    \item \textbf{A large-scale macOS CUA benchmark} of 421 high-quality tasks, combining manually ported OSWorld tasks with verified macOSWorld tasks and 49 new macOS-specific tasks into a unified evaluation suite.
    \item \textbf{Human verification of all tasks}, ensuring each task is executable, unambiguous, and correctly specified, providing a higher-quality signal than automated task generation or partial review.
    \item \textbf{Full reproducibility}, with all code publicly released to enable community extension of the benchmark.
    \item \textbf{Evaluation of models}, establishing baseline results and surfacing strengths and limitations of existing CUAs on macOS.
\end{itemize}

Our evaluation establishes baseline results for current CUAs on macOS and reveals a consistent pattern: performance degrades across all evaluated models relative to Linux, suggesting macOS poses a genuinely harder environment for current GUI agents.

\section{Related Work}

\begin{table*}[htbp]
\centering
\caption{Comparison of CUA benchmarks. \cmark~indicates the property is present, \xmark~that it is absent, and ---~that it is not applicable or not reported.}
\label{tab:benchmark_comparison}
\begin{tabular}{lllccc}
\toprule
\textbf{Benchmark} & \textbf{Platform} & \textbf{\# Tasks} & \textbf{\# Apps} & \textbf{3rd-party Apps} & \textbf{Manual Verif.} \\
\midrule
\multicolumn{6}{l}{\textit{Offline benchmarks}}\\
\midrule
Mind2Web~\cite{deng2023mind2web}                      & Web             & 2,350 & 137   & ---    & \xmark \\
AITW~\cite{rawles2023androidwildlargescaledataset}    & Android         & 30k   & 159   & ---    & \xmark \\
ScreenSpot~\cite{cheng2024seeclick}                   & Multi           & 1,200 & ---   & \cmark & \cmark \\
ScreenSpot-V2~\cite{wu2024atlas}                      & Multi           & 1,272 & ---   & \cmark & \cmark \\
ScreenSpot-Pro~\cite{li2025screenspotpro}             & Multi           & 1,581 & 23    & \cmark & \cmark \\
GUIrilla-Gold~\cite{garkot2026guirillascalableframeworkautomated} & macOS    & 1,283 & 219   & \cmark & \cmark \\
\midrule
\multicolumn{6}{l}{\textit{Online benchmarks}}\\
\midrule
OSWorld~\cite{OSWorld}                        & Linux, Win      & 369 & 9   & \cmark & \cmark \\
WAA~\cite{bonatti2024windows}                 & Windows         & 154 & 15  & \cmark & \xmark \\
macOSWorld~\cite{macosworld}                  & macOS           & 202 & 30  & \xmark & \xmark \\
WebArena~\cite{zhou2024webarena}              & Web             & 812 & --- & ---    & \xmark \\
VisualWebArena~\cite{koh2024visualwebarena}   & Web             & 910 & --- & ---    & \xmark \\
WorkArena~\cite{workarena2024}                & Web             &  33 & 1   & ---    & \xmark \\
AndroidWorld~\cite{rawles2024androidworld}    & Android         & 116 & 20  & \cmark & \xmark \\
B-MoCA~\cite{lee2024benchmarking}             & Android         & 131 & 10  & \cmark & \xmark \\
\midrule
\textbf{MacArena (ours)}                      & \textbf{macOS} & \textbf{421} & 50 & \cmark & \cmark \\
\bottomrule
\end{tabular}
\end{table*}

\subsection{GUI Agents and Computer-Use Systems}
Early GUI agents were built around prompt-based pipelines that combined frontier vision-language models with modular planning and memory components. Systems such as UFO~\cite{zhang2025ufo} and SeeAct~\cite{zheng2023seeact} demonstrated that GPT-4V could complete desktop and web tasks by reasoning over screenshots, while multi-agent frameworks decomposed tasks into subtasks handled by specialized subagents. Though effective in constrained settings, these systems were limited by the capabilities of the underlying models and the overhead of multi-step orchestration.

A second line of work focused on improving visual grounding, the ability to precisely localize UI elements from natural language descriptions. CogAgent~\cite{hong2023cogagent} introduced a dual-encoder architecture trained specifically on GUI layouts, and subsequent models developed dedicated grounding modules that allowed agents to click and interact with greater spatial precision. This grounding capability became a prerequisite for reliable performance on complex desktop tasks.

More recently, end-to-end trained agents have largely superseded prompt-based pipelines. Models such as UI-TARS~\cite{qin2025ui} and Aguvis~\cite{xu2024aguvis} are trained natively on large collections of GUI interaction trajectories across desktop, web, and mobile platforms, achieving strong generalization without relying on external orchestration. These single-model agents are simpler to deploy and have become the dominant paradigm for GUI agent development.

Reinforcement learning has emerged as a further lever for improving agent performance. DigiRL~\cite{bai2024digirl} demonstrated offline-to-online RL fine-tuning on Android tasks, and ComputerRL~\cite{lai2025computerrl} scaled online RL to desktop environments using thousands of parallel virtual machines. UI-TARS-2~\cite{wang2025uitars2technicalreportadvancing} extended this with a multi-turn RL framework that generates training trajectories at scale. Across all of this work, OSWorld and AndroidWorld have served as the primary training and evaluation environments, underscoring the absence of a comparable macOS benchmark for desktop GUI research.

\subsection{Computer Use Benchmarks}

Before interactive benchmarks, researchers developed offline datasets to evaluate agents on static screenshots. Mind2Web~\cite{deng2023mind2web} and AITW~\cite{rawles2023androidwildlargescaledataset} collected large sets of human demonstrations for web and mobile navigation, enabling supervised training of GUI policies. ScreenSpot~\cite{cheng2024seeclick}, ScreenSpot-V2~\cite{wu2024atlas}, and ScreenSpot-Pro~\cite{li2025screenspotpro} established benchmarks focused specifically on element localization, the ability to map a natural language instruction to the correct location on a screen. GUIrilla~\cite{garkot2026guirillascalableframeworkautomated} extended this to macOS, providing localization annotations across a wide range of third-party applications. While these offline benchmarks remain useful for model development, they do not capture the sequential decision-making, error recovery, and dynamic environment feedback that define real-world agent behavior, and therefore do not measure whether an agent can complete tasks in a live environment.

OSWorld~\cite{OSWorld} is the most comprehensive existing interactive benchmark, covering Linux and Windows with real applications, multi-step tasks, and automated grading based on functional outcomes. Tasks are initialized from virtual machine snapshots. OSWorld has become the standard environment for both training and evaluating desktop GUI agents. For macOS specifically, macOSWorld~\cite{macosworld} introduced a benchmark targeting Apple's built-in applications, such as Finder, Safari, and Calendar. Yet its tasks tend to be simpler, and more narrowly defined than those in OSWorld, with coverage limited almost entirely to first-party software.

Hardware compatibility compounds this limitation. Following Apple's 2020 transition away from Intel processors, x86-based macOS environments are increasingly misaligned with real-world usage: Apple Silicon now powers the entire Mac lineup, and Intel-based machines are no longer manufactured. Yet macOSWorld was designed around x86 virtual machines with no native Apple Silicon support. While cloud-based evaluation on Apple Silicon hardware (e.g., via EC2 Mac instances) is technically possible, it introduces significant cost overhead that makes large-scale benchmarking and RL training pipelines impractical.

MacArena addresses these gaps directly. macOS presents distinct challenges for GUI agents: from its application conventions and complex window management to the widespread use of third-party software, that existing benchmarks leave largely unexamined. As shown in Table~\ref{tab:benchmark_comparison}, MacArena is the only macOS online benchmark combining third-party application coverage with full manual verification.

\section{MacArena Environment}
\subsection{Problem Formulation}

\begin{table*}[htbp]
\centering
\caption{Supported actions and their parameters.}
\label{tab:action_space}
\renewcommand{\arraystretch}{1.2}
\begin{tabular}{lllll}
\toprule
\textbf{Category} & \textbf{Action} & \textbf{Params} & \textbf{Description} \\
\midrule
\multirow{8}{*}{\textbf{Mouse}}
  & \texttt{MOVE\_TO}      & \texttt{x}, \texttt{y}   & Move cursor to position \\
  & \texttt{CLICK}         & \texttt{x}, \texttt{y}, \texttt{button} (l/r/m)   & Click at position \\
  & \texttt{RIGHT\_CLICK}  &  \texttt{x}, \texttt{y}  & Right-click at position \\
  & \texttt{DOUBLE\_CLICK} &  \texttt{x}, \texttt{y}  & Double-click at position \\
  & \texttt{DRAG\_TO}      & \texttt{x}, \texttt{y}   & Drag to target position \\
  & \texttt{SCROLL}        & \texttt{dx}, \texttt{dy} & Scroll by delta \\
  & \texttt{MOUSE\_DOWN}   & \texttt{button}          & Press and hold mouse button \\
  & \texttt{MOUSE\_UP}     & \texttt{button}          & Release mouse button \\
\midrule
\multirow{5}{*}{\textbf{Keyboard}}
  & \texttt{TYPING}    & \texttt{text}     & Type a sequence of characters \\
  & \texttt{PRESS}     & \texttt{key}      & Press a single key \\
  & \texttt{KEY\_DOWN} & \texttt{key}      & Hold a key down \\
  & \texttt{KEY\_UP}   & \texttt{key}      & Release a held key \\
  & \texttt{HOTKEY}    & \texttt{[keys]} & Press a key combination \\
\midrule
\multirow{3}{*}{\textbf{Terminal}}
  & \texttt{WAIT} & — & Sleep until next action \\
  & \texttt{FAIL} & — & Signal task cannot be completed \\
  & \texttt{DONE} & — & Signal task successfully completed \\
\bottomrule
\end{tabular}
\end{table*}

We formalize an autonomous agent interaction in MacArena as a Partially Observable Markov Decision Process (POMDP)~\cite{OSWorld, bonatti2024windows, macosworld}, defined by the tuple $(\mathcal{S}, \mathcal{O}, \mathcal{A}, \mathcal{T}, \Omega, r, \gamma, \mu_0, \mathcal{G}, p_g, \varphi)$, 
where $\mathcal{S}$ is the full state space of the macOS environment (including hidden system states such as background processes and file system contents), $\mathcal{O}$ is the observation space accessible to the agent (e.g., screenshots and accessibility trees), $\mathcal{A}$ is the action space of mouse and keyboard interactions (full action space is listed in Table~\ref{tab:action_space}). $\mathcal{T}: \mathcal{S} \times \mathcal{A} \rightarrow \mathcal{S}$ is the deterministic transition function, $\Omega$ is the observation function mapping states to observations, $r: \mathcal{S} \times \mathcal{A} \times \mathcal{G} \rightarrow \mathbb{R}$ is the reward function, $\gamma$ is the discount factor, $\mu_0$ is the initial state distribution, $\mathcal{G}$ is the space of task goals (expressed as natural language instructions), $p_g$ is the distribution over goals, and $\varphi: \mathcal{O} \rightarrow \mathcal{G}$ is a mapping from observations to goals.

At each timestep $t$, the agent receives an observation $o_t \in \mathcal{O}$ consisting of a screenshot of the current macOS desktop, optionally with the accessibility tree. Accessibility (a11y) is a structured, hierarchical representation of UI elements (buttons, text fields, menus) exposed by macOS via the Accessibility API, providing element labels, roles, and bounding boxes without requiring visual parsing. Based on this observation, the agent produces an executable action $a_t \in \mathcal{A}$, e.g., click. The action is executed within the virtual machine, transitioning the environment to a new state $s_{t+1} \in \mathcal{S}$ and yielding a new observation $o_{t+1} \in \mathcal{O}$. This loop continues until the agent emits a terminal action (\texttt{DONE} or \texttt{FAIL}) or the maximum number of steps is reached.

MacArena implements an execution-based reward function $r: \mathcal{S} \times \mathcal{A} \times \mathcal{G} \rightarrow [0, 1]$. At the final step, a custom evaluation script compares the resulting environment state to the task objective and assigns a score, with higher values indicating greater task completion.

\subsection{Benchmark Structure}
\label{sec:benchmark_structure}

MacArena consists of three core components: the environment, the tasks, and the evaluation framework.

\paragraph{Environment.}
MacArena runs inside virtual machines (VMs) managed by UTM\footnote{\url{https://github.com/utmapp/UTM}}, which is built on Apple's native Virtualization framework (see Appendix~\ref{sec:apple_virtualization} for details). We maintain two distinct VMs. The first is dedicated to tasks sourced from OSWorld and macOSWorld; it was configured manually to satisfy the setup assumptions of those benchmarks, including installed applications, system permissions, and verified evaluation conditions. The second VM is purpose-built for MacArena's own tasks and is provisioned entirely via an automated build script. This approach eliminates manual configuration, simplifies migration across macOS versions, and makes it straightforward to add new applications in the future.

Since UTM does not natively support VM snapshot revert, we adopt a copy-on-use strategy: before each task episode, the original VM image is copied to a temporary instance used for the episode and discarded upon completion. This guarantees a clean, reproducible initial state for every evaluation run.

The environment exposes two types of observations to the agent: pixel-level screenshots and, optionally, the macOS accessibility tree taken using \texttt{macapptree}~\cite{garkot2026guirillascalableframeworkautomated}. 

\paragraph{Tasks.}
MacArena contains 421 tasks organized into two dimensions: application and task type, including cross-application workflows that require coordinating multiple macOS apps. Tasks are sourced from two existing benchmarks: OSWorld and macOSWorld, and supplemented with our own newly collected tasks, all reviewed by humans to ensure each task is executable, unambiguous, and correctly specified.

Each task is defined by three required components. The \texttt{instruction} field provides a natural language description of what the agent must accomplish. The \texttt{pre\_command/config} field specifies an initialization procedure that prepares the VM for the task, such as downloading required files, launching applications, or opening documents. The \texttt{evaluator} field defines a deterministic function that programmatically verifies whether the task was completed successfully by inspecting the final environment state.

MacArena supports two task formats inherited from its source benchmarks. The OSWorld format uses a set of predefined evaluation functions composed via a structured configuration file. The macOSWorld format uses shell scripts for both initialization and evaluation, offering greater flexibility for tasks that require more complex or platform-specific logic. Our own tasks follow both formats.

\paragraph{Evaluation Framework.}
Task success in MacArena is determined by execution-based evaluation: after the agent emits a terminal action, the corresponding evaluator script is executed against the final VM state. Each evaluator returns a value in $[0, 1]$ indicating how well the task was completed; the higher, the better. Evaluation criteria vary by task and may inspect file contents, application state, system properties, or the output of shell commands, depending on what the task requires.

Each of the 49 tasks in the MacArena subset is associated with a unique, hand-crafted evaluation script, yielding 49 distinct evaluation functions in total. This one-to-one correspondence between tasks and evaluators reflects the diversity of verification requirements
across macOS applications and task types. 

\subsection{Benchmark Statistics}

MacArena comprises 421 tasks in total, drawn from three sources: 221 tasks adapted from OSWorld, 151 tasks from macOSWorld, and 49 newly collected tasks across 5 categories for the MacArena-specific subset. The MacArena-specific subset spans 20 macOS applications. OSWorld and macOSWorld tasks retain their original application coverage, which includes both macOS-exclusive apps and cross-platform productivity tools.

Tasks are organized into 20 categories reflecting the structure of their source benchmarks. The OSWorld-derived tasks cover seven categories: \texttt{chrome}, \texttt{gimp}, \texttt{libreoffice\_calc}, \texttt{libreoffice\_writer}, \texttt{thunderbird}, \texttt{vs\_code}, and \texttt{multi\_apps}. The macOSWorld-derived tasks span seven categories: \texttt{sys\_and\_interface}, \texttt{sys apps}, \texttt{file management}, \texttt{productivity}, \texttt{media}, \texttt{multi apps}, and \texttt{advanced}. The MacArena-specific tasks are organized into 5 categories, inheriting macOSWorld names: \texttt{file management}, \texttt{system and interface}, \texttt{advanced apps}, \texttt{built-in apps}, and \texttt{productivity}, covering tasks that require interaction with a single macOS application at a time.

\begin{figure}[htbp]
    \centering
    \includegraphics[width=\linewidth]{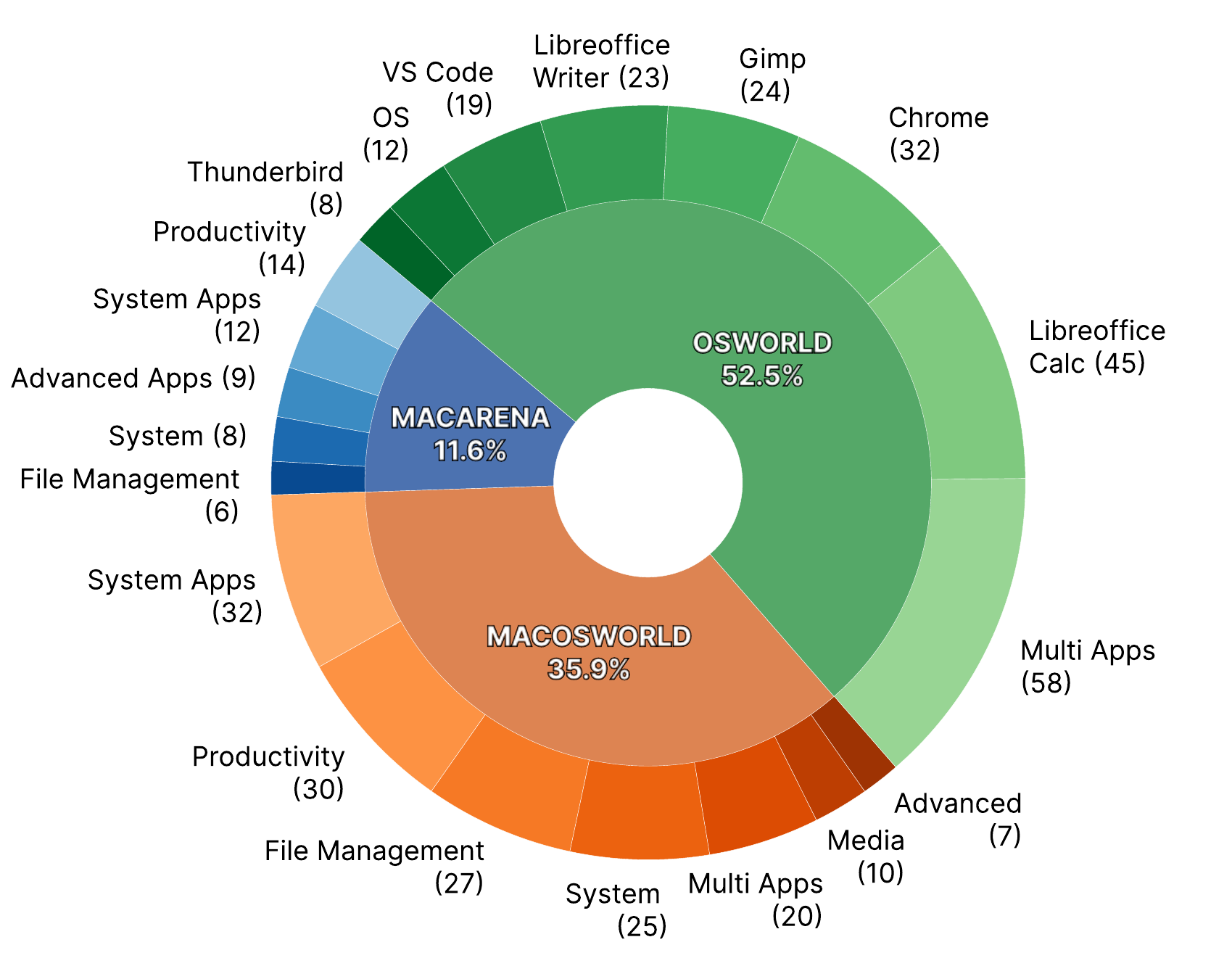}
    \caption{Distribution of tasks per category across the full MacArena benchmark.}
    \label{fig:distribution}
\end{figure}

Figure~\ref{fig:distribution} shows the distribution of tasks across all 20 categories, illustrating the diversity of macOS use cases covered by MacArena. 

\section{Experiments}
\begin{table*}[t]
\centering
\caption{Success rates (\%) of baseline agents on MacArena.
Best result in each row is \textbf{bold}.}
\label{tab:main_results}
\setlength{\tabcolsep}{5pt}
\begin{tabular}{lcccc}
\toprule
\textbf{Category} &
\textbf{UI-TARS-1.5 7B} &
\textbf{Qwen3-VL 2B} &
\textbf{Qwen3-VL 4B} &
\textbf{OpenAI CUA} \\
\midrule

\multicolumn{5}{l}{\textit{OSWorld subset}} \\
\midrule
\quad Overall            & \textbf{21.27} & 9.95  & 16.36          & 16.74         \\
\quad Chrome             & 28.12 & 15.62 & \textbf{37.50} & \textbf{37.50} \\
\quad GIMP               & \textbf{41.67} & 8.33  & 12.50 & 4.17  \\
\quad LibreOffice Calc   & \textbf{6.67}  & 2.22  & 4.44           & 4.44  \\
\quad LibreOffice Writer & 17.39 & 13.04 & 17.39          & \textbf{21.74} \\
\quad Multi-App          & 0.00  & 1.72  & \textbf{3.45}  & 1.72           \\
\quad OS                 & \textbf{50.00} & 16.67 & 36.36 & 33.33 \\
\quad Thunderbird        & 25.00 & 25.00 & 25.00          & \textbf{37.50} \\
\quad VS Code            & \textbf{68.42} & 31.58 & 36.84 & 47.37 \\
\midrule

\multicolumn{5}{l}{\textit{macOSWorld subset}} \\
\midrule
\quad Overall            & 24.50 & 15.89 & 39.74          & \textbf{52.32} \\
\quad Advanced           & 0.00  & 14.29 & 14.29          & \textbf{42.86} \\
\quad File Management    & 25.93 & 7.41  & 37.04          & \textbf{40.74} \\
\quad Media              & \textbf{40.00} & 20.00 & 30.00 & 30.00 \\
\quad Multi-App          & 5.00  & 10.00 & \textbf{15.00} & \textbf{15.00} \\
\quad Productivity       & 26.67 & 20.00 & 63.33          & \textbf{70.00} \\
\quad Sys \& Interface   & 32.00 & 32.00 & 44.00          & \textbf{60.00} \\
\quad System Apps        & 28.12 & 9.38  & 40.62          & \textbf{71.88} \\
\midrule

\multicolumn{5}{l}{\textit{MacArena subset}} \\
\midrule
\quad Overall          & 10.20 & 4.08  & 12.24          & \textbf{36.73} \\
\quad Productivity     & 7.14  & 7.14  & 14.29          & \textbf{28.57} \\
\quad Sys \& Interface & 25.00 & 0.00  & 12.50          & \textbf{37.50} \\
\quad File Management  & 0.00  & 0.00  & 0.00           & 0.00 \\
\quad Advanced         & 11.11 & 11.11 & 11.11          & \textbf{55.56} \\
\quad System Apps      & 8.33  & 0.00  & 16.67          & \textbf{50.00} \\
\midrule
\textit{Overall} & 21.14 & 11.40 & 24.23 & \textbf{31.83} \\
\bottomrule
\end{tabular}
\end{table*}

We evaluate four baseline agents on MacArena: UI-TARS-1.5 7B~\cite{qin2025ui}, Qwen3-VL 2B, Qwen3-VL 4B~\cite{qwen3technicalreport}, and OpenAI Computer Use Preview~\cite{cua2025}. All agents interact with the macOS virtual machine through raw mouse and keyboard actions. Each agent receives a screenshot of the current desktop at every step. Each task is limited to 15 steps, and each model has 2 runs. Task success is determined by execution-based evaluation scripts as described in Section~\ref{sec:benchmark_structure}. We report Success Rate~(SR) as the primary metric, defined as the percentage of tasks for which the evaluation script returns a positive result.

Table~\ref{tab:main_results} reports the success rates of all four agents across the three subsets and 16 task categories. OpenAI Computer Use Preview achieves the highest overall success rate of 31.83\%, followed by Qwen3-VL 4B (24.23\%), UI-TARS-1.5 7B (21.14\%), and Qwen3-VL 2B (11.40\%).

\subsection{macOS vs.\ Linux Performance Gap}

A key motivation for MacArena is the hypothesis that macOS presents distinct challenges for GUI agents beyond what existing Linux-based benchmarks capture. To investigate this, we compare model performance on the OSWorld subset of MacArena against officially reported scores on the original OSWorld benchmark, which runs on Ubuntu Linux with a 15-step budget. Table~\ref{tab:osworld_comparison} presents this comparison for models where official scores are publicly available.

\begin{table}[htbp]
\centering
\caption{Success rates (\%) on the original OSWorld benchmark (Ubuntu, 15 steps) vs. the OSWorld subset of MacArena (macOS, 15 steps).}
\label{tab:osworld_comparison}
\begin{tabular}{lccc}
\toprule
\textbf{Model} &
\textbf{Ubuntu} &
\textbf{macOS} &
\textbf{$\Delta$} \\
\midrule
UI-TARS-1.5 7B  & 24.5 & 21.27 & $-$3.23 \\
OpenAI CUA      & 26.0 & 16.74 & $-$9.26 \\
Qwen3-VL 2B     & 17.0 &  9.95 & $-$7.05 \\
Qwen3-VL 4B     & 26.2 & 16.36 & $-$9.84 \\
\bottomrule
\end{tabular}
\end{table}

Both models with available reference scores show a meaningful drop when evaluated on macOS, despite the task set being identical. This gap is due to macOS introducing platform-specific differences in application appearance, keyboard shortcuts, window management, and system behavior that models trained primarily on Linux and Windows trajectories are not adapted to. The consistent degradation across both models suggests that macOS poses a genuinely harder environment for current GUI agents.

\subsection{Analysis}
\label{sec:analysis}

\paragraph{Multi-app tasks.} The multi-app tasks category involves interaction between 2 or more applications. This category remains the hardest category across all models and both the OSWorld and macOSWorld subsets, with most agents scoring at or near 0\%. This is consistent with prior work~\cite{OSWorld, macosworld} and indicates that coordinating across applications remains an open challenge even for state-of-the-art models.

\paragraph{Category-level strengths and weaknesses.} UI-TARS-1.5 7B achieves the highest scores on VS Code (68.42\%) and OS tasks (50.00\%) within the OSWorld subset, suggesting stronger adaptation to terminal and code editing workflows. In contrast, it performs poorly on macOS-native categories such as Advanced (0.00\%) and MacArena (10.2\%), indicating limited exposure to macOS-specific application patterns during training. OpenAI CUA dominates the macOSWorld subset, particularly System Apps (71.88\%) and Productivity (70.00\%), reflecting stronger adaptation to native macOS applications.

\paragraph{Divergence between OSWorld and MacArena subsets.} An interesting reversal emerges when comparing performance on the OSWorld subset against the MacArena-specific subset. UI-TARS-1.5 7B outperforms OpenAI CUA on the OSWorld subset (21.27\% vs.\ 16.74\%), yet this advantage completely inverts on the MacArena-specific tasks, where OpenAI CUA scores 36.73\% against UI-TARS-1.5 7B's 10.2\%, a gap of over 26.5 percentage points in the opposite direction. This divergence suggests that strong performance on tasks originally designed for Linux does not transfer to novel macOS-native tasks. UI-TARS-1.5 7B likely benefits from having seen OSWorld-style tasks or similar Linux-based GUI trajectories during training, which gives it an advantage on the OSWorld subset despite the platform shift. However, when faced with genuinely new macOS applications and interaction patterns in the MacArena subset, this advantage disappears entirely. OpenAI CUA, by contrast, appears to have broader macOS-specific knowledge, possibly from training on diverse real-world computer use data that includes macOS. This result highlights an important limitation of evaluating GUI agents exclusively on existing benchmarks: a model can appear competitive by pattern-matching previously seen task structures while failing to generalize to new environments. MacArena's own task subset is specifically designed to surface this gap.

\paragraph{Task difficulty via step consumption.}
To further characterize the relative difficulty of each subset, we analyze the average number of steps consumed per task by OpenAI CUA, the best-performing model in our evaluation. We report both the average steps across all tasks and the average steps on completed tasks only, as the latter reflects the true complexity of tasks the model was capable of solving. Table~\ref{tab:avg_steps} summarizes these statistics aggregated by subset.

\begin{table}[htbp]
\centering
\caption{Average steps consumed by OpenAI CUA per subset, across all tasks, and only on completed tasks.}
\label{tab:avg_steps}
\begin{tabular}{lcc}
\toprule
\textbf{Subset} &
\textbf{Avg.\ Steps (All)} &
\textbf{Avg.\ Steps (Done)} \\
\midrule
macOSWorld & 10.92 & 8.05  \\
OSWorld    & 13.88 & 11.08 \\
MacArena   & 13.96 & 12.69 \\
\bottomrule
\end{tabular}
\end{table}

The macOSWorld subset has the lowest average step consumption across all tasks (10.92) and on completed tasks (8.05), indicating that its tasks are, on average, shorter and less complex than those in the other two subsets. This provides a quantitative explanation for why models achieve higher success rates on macOSWorld: beyond any potential platform familiarity, the tasks themselves require fewer interaction steps to complete. The OSWorld subset requires more steps on average (13.88 overall, 11.08 for completed tasks). The MacArena-specific subset exhibits the highest step consumption of all three, both overall (13.96) and on completed tasks (12.69).

\section{Limitations and Future Work}

\paragraph{Automatic task generation.}
All tasks in MacArena were manually created by human annotators, which is
time-consuming and limits scalability. A promising direction for future work
is to automate task generation by leveraging LLMs to synthesize diverse,
plausible task instructions, potentially guided by
application-specific schemas or interaction logs. However, automatically
generated tasks introduce quality concerns: instructions may be ambiguous,
infeasible, or trivially easy. Ensuring that generated tasks are valid,
non-redundant, and appropriately challenging would therefore require either
human validation or automated feasibility checks, for example, by verifying
that a reference agent can complete the task within a bounded number of
steps. Addressing this pipeline would substantially reduce annotation cost
and enable MacArena to scale to a broader range of applications and task
types.

\paragraph{Human performance baseline.}
While tasks in MacArena are 100\% human-verified and each task is possible to do, benchmark does not currently include a human performance study. Establishing human baselines would provide an important reference point for interpreting model results and understanding the remaining headroom for future improvement, as done in OSWorld~\cite{OSWorld} and macOSWorld~\cite{macosworld}.

\paragraph{Use of LLMs.}
Parts of this work were refined using LLMs for better readability and structure.

\section{Conclusion}
We presented MacArena, a benchmark for evaluating GUI agents on macOS, comprising 421 tasks drawn from OSWorld and macOSWorld, as well as a newly collected set of 49 macOS-native tasks spanning 20 applications. MacArena provides a unified evaluation framework that runs all tasks within a reproducible virtual machine environment using execution-based evaluation scripts, enabling direct comparison of agent behavior across task origins on a single platform.

Our evaluation of agents reveals that macOS remains a challenging and underexplored environment for current GUI agents. All models with available reference scores perform worse on the same task set when evaluated on macOS compared to the original Linux-based OSWorld environment, confirming that the platform gap is real. Furthermore, relative model rankings shift substantially depending on which subset of MacArena is examined: a model that leads on OSWorld-derived tasks can fall far behind on novel macOS-native tasks. This suggests that strong performance on existing benchmarks may reflect familiarity with specific task distributions rather than genuine cross-platform GUI understanding, and that macOS-native tasks are necessary to expose this limitation.

MacArena is designed to serve as a foundation for GUI agents that generalize reliably across operating systems, and to establish macOS as a first-class evaluation target alongside Linux and Windows.

\section*{Impact Statement}
This paper introduces a benchmark for evaluating computer-use agents on macOS. The primary goal is to advance the scientific study of GUI agents by providing a more comprehensive and reproducible evaluation environment. We do not anticipate direct negative societal consequences specific to this work. Improvements in agent capabilities enabled by better evaluation tools may contribute to automating knowledge work, with socioeconomic implications widely discussed in the AI community. We do not believe these consequences require specific highlighting here beyond what is already widely discussed in the context of autonomous agent research.

\section*{Acknowledgements}
We thank the Armed Forces of Ukraine for their courage and sacrifice, which enabled us to complete this work.

We also thank Mariya Hirna for her support in coordinating and managing project logistics, as well as for many meaningful discussions that helped shape this work. We are grateful to Bohdan Antoniuk for his assistance with virtual machine setup and scaling.


\bibliography{main}
\bibliographystyle{icml2026}

\newpage
\appendix
\onecolumn
\section{Apple Virtualization Framework}
\label{sec:apple_virtualization}

MacArena is built on Apple's Virtualization framework\footnote{\url{https://developer.apple.com/documentation/virtualization}}, a native hypervisor API introduced in macOS 11 (Big Sur) that enables the creation and management of virtual machines directly on Apple Silicon hardware. Unlike traditional virtualization solutions such as QEMU, which rely on software emulation, the framework leverages the hardware virtualization extensions of M-series chips to run guest operating systems at near-native performance. 

A key practical constraint of the framework is that Apple Silicon hosts can run at most two macOS guest VMs simultaneously — a hard limit enforced at the system level. This restricts the degree of parallelism available per machine and is an important consideration for benchmark throughput. In MacArena, we account for this by running up to two evaluation tasks in parallel on a single host. To scale evaluation further, multiple host machines can be employed independently, as the per-host VM limit does not preclude distributed execution across several Apple Silicon machines.

\end{document}